# Neural Sentinel: Unified Vision Language Model (VLM) for License Plate Recognition with Human-in-the-Loop Continual Learning

**Karthik Sivakoti**

karthiksivakoti@utexas.edu

The University of Texas at Austin, Masters in AI, Department of CS

**Abstract**

Traditional Automatic License Plate Recognition (ALPR) systems employ multi-stage pipelines consisting of object detection networks followed by separate Optical Character Recognition (OCR) modules, introducing compounding errors, increased latency, and architectural complexity. This research presents Neural Sentinel, a novel unified approach that leverages Vision Language Models (VLMs) to perform license plate recognition, state classification, and vehicle attribute extraction through a single forward pass. Our primary contribution lies in demonstrating that a fine-tuned PaliGemma 3B model, adapted via Low-Rank Adaptation (LoRA), can simultaneously answer multiple visual questions about vehicle images, achieving 92.3% plate recognition accuracy, which is a 14.1% improvement over EasyOCR and 9.9% improvement over PaddleOCR baselines. We introduce a Human-in-the-Loop (HITL) continual learning framework that incorporates user corrections while preventing catastrophic forgetting through experience replay, maintaining a 70:30 ratio of original training data to correction samples. The system achieves a mean inference latency of 152ms with an Expected Calibration Error (ECE) of 0.048, indicating well calibrated confidence estimates. Additionally, the VLM first architecture enables zero-shot generalization to auxiliary tasks including vehicle color detection (89%), seatbelt detection (82%), and occupancy counting (78%) without task specific training. Through extensive experimentation on real world toll plaza imagery, we demonstrate that unified vision language approaches represent a paradigm shift in ALPR systems, offering superior accuracy, reduced architectural complexity, and emergent multi-task capabilities that traditional pipeline approaches cannot achieve.

## 1. Introduction

License plate recognition constitutes a fundamental component of intelligent transportation systems, enabling applications ranging from automated toll collection and parking management to law enforcement and traffic monitoring. The global ALPR market, valued at approximately $4.13 billion in 2026, continues to expand as transportation infrastructure increasingly relies on automated vehicle identification (Research Nester, 2026). Despite decades of research and commercial deployment, achieving robust plate recognition across diverse environmental conditions, plate formats, and vehicle types remains a significant technical challenge.

### 1.1 Background and Motivation

The predominant approach to ALPR has historically followed a sequential multi-stage pipeline architecture. Traditional systems employ a cascade of specialized components: (1) a vehicle detection module, typically based on convolutional neural networks such as YOLO (Redmon et al., 2016) or Faster R-CNN (Ren et al., 2015), identifies vehicle regions within the input frame; (2) a plate localization network extracts the license plate region of interest; (3) character segmentation algorithms isolate individual characters; and (4) an OCR engine, such as Tesseract (Smith, 2007) or commercial alternatives, performs character recognition.

This pipeline architecture, while conceptually straightforward, introduces several fundamental limitations that motivate our research:

- **Error Propagation**: Each stage in the pipeline can introduce errors that compound through subsequent stages. A missed vehicle detection results in complete plate recognition failure; imprecise plate localization degrades OCR accuracy; and character segmentation errors lead to incorrect readings. Silva & Jung (2020) demonstrated that pipeline error rates can exceed the sum of individual stage error rates due to this cascading effect.
- **Semantic Disconnection**: Traditional pipelines treat each stage as an independent optimization problem, failing to leverage the semantic relationships between visual elements. The OCR module processes cropped plate regions without access to contextual information about the vehicle type, viewing angle, or environmental conditions that could inform character disambiguation.
- **Computational Redundancy**: Multi-stage architectures require multiple forward passes through separate neural networks, each with independent feature extraction. This redundancy increases both computational cost and inference latency, limiting real time deployment scenarios.
- **Limited Generalization**: Pipeline components trained for specific tasks (e.g., plate detection) cannot trivially extend to related tasks (e.g., vehicle make/model recognition) without additional specialized modules, resulting in architectural bloat as system requirements expand.

## 1.2 The Vision Language Model Paradigm

Recent advances in Vision Language Models (VLMs) have demonstrated remarkable capabilities in unified visual understanding tasks. Models such as CLIP (Radford et al., 2021), BLIP-2 (Li et al., 2023), LLaVA (Liu et al., 2023), and PaliGemma (Beyer et al., 2024) learn joint representations of visual and textual information through large-scale pretraining, enabling them to answer arbitrary questions about image content through natural language interfaces.

This paradigm shift from task specific architectures to general purpose visual reasoning systems offers compelling advantages for ALPR applications. Central to this is Unified Representation Learning, where VLMs encode visual information into rich semantic representations that capture object relationships, textual content, and contextual attributes simultaneously; a single forward pass through the vision encoder produces features suitable for multiple downstream tasks. Additionally, these models utilize a Natural Language Interface, accepting questions in natural language and generate corresponding answers without the need for specialized output heads. This flexibility allows for rapid task adaptation without architectural modifications. Finally, pretrained VLMs offer robust Transfer Learning and Zero-Shot Capabilities, encoding extensive world knowledge that allows for generalization to tasks not explicitly present in fine-tuning data, such as vehicle color identification or damage detection, without explicit supervision.

## 1.3 Challenges in Applying VLMs to ALPR

Despite these advantages, directly applying pretrained VLMs to ALPR presents several challenges that our work addresses. First, there is a significant domain gap, as general purpose VLMs are pretrained on diverse internet imagery that differs substantially from the specific conditions of toll plaza and traffic camera footage; fine grained character recognition on license plates requires specialized adaptation. Furthermore, fine tuning VLMs on this domain specific data risks catastrophic forgetting, effectively degrading performance on capabilities acquired during pretraining. Therefore, maintaining zero-shot generalization while improving ALPR accuracy requires careful training strategies.

Operational challenges also persist. Deployment constraints in production ALPR systems require consistent, low latency inference with calibrated confidence estimates. Since many VLM architectures prioritize capability over efficiency, careful model selection and optimization are necessary for real time deployment. Additionally, real world deployments encounter edge cases and distribution shifts that degrade performance over time, necessitating mechanisms for continuous improvement and learning from operational feedback without requiring complete retraining.

## 1.4 Research Contributions

This paper introduces Neural Sentinel, a comprehensive framework for VLM based license plate recognition with human-in-the-loop continual learning. Our principal contributions begin with a VLM First Architecture, where we demonstrate that a unified vision language approach using PaliGemma 3B outperforms traditional multi-stage pipelines. This approach achieves 92.3% accuracy compared to baselines like EasyOCR (78.2%) and PaddleOCR (82.4%), representing a paradigm shift in ALPR system design. We further formulate ALPR as a Multi-Task Visual Question Answering problem, simultaneously extracting plate numbers, state classifications, vehicle make/model, and color attributes through parallel question answering, which reduces inference latency by 43% compared to sequential pipeline approaches.

To support long-term viability, we introduce Experience Replay for Continual Learning, a human-in-the-loop training framework that incorporates user corrections while preventing catastrophic forgetting by maintaining a 70:30 ratio of original training data to correction samples. We pair this with Parameter-Efficient Adaptation (LoRA) targeting attention projection matrices and feed-forward layers, achieving competitive performance with only 0.1% trainable parameters relative to the base model.

Finally, we validate the system's reliability and versatility through rigorous testing. We provide a Confidence Calibration Analysis, demonstrating an Expected Calibration Error (ECE) of 0.048, which indicates that the model's confidence scores reliably reflect actual prediction accuracy. Furthermore, our Zero-Shot Auxiliary Task Evaluation assesses emergent capabilities on tasks not explicitly present in the fine-tuning data; the model achieves 82% accuracy in seatbelt detection, 78% in occupancy counting, and 89% in vehicle color identification, demonstrating the robust generalization benefits of the VLM approach.

## 2. Related Work

### 2.1 Traditional ALPR Systems

Automatic License Plate Recognition has evolved through several technological generations over the past three decades. Early systems employed classical computer vision techniques including edge detection, morphological operations, and template matching for character recognition (Anagnostopoulos et al., 2008). These approaches achieved moderate success under controlled conditions but exhibited significant sensitivity to illumination changes, plate orientation, and image quality.

The introduction of deep learning transformed ALPR performance. Convolutional neural networks replaced handcrafted features for both plate detection and character recognition. Silva & Jung (2020) presented a two-stage approach using YOLO for plate detection followed by a CNN based OCR module, achieving 93.5% accuracy on Brazilian plates under favorable conditions. Laroca et al. (2018) introduced an end-to-end trainable network that jointly optimized detection and recognition, reducing error propagation between stages.

More recent work has explored attention mechanisms and transformer architectures within the pipeline paradigm. Wang et al. (2022) incorporated spatial attention for improved plate localization, while Zhang et al. (2023) applied transformer encoders for sequence-to-sequence character recognition. Despite these advances, the fundamental multi-stage architecture remains predominant, with its associated limitations in error propagation and computational redundancy.

### 2.2 Vision Language Models

Vision Language Models represent a convergence of advances in computer vision and natural language processing. The CLIP model (Radford et al., 2021) demonstrated that contrastive pretraining on 400 million image-text pairs yields representations with remarkable zero-shot transfer capabilities. CLIP's dual-encoder architecture separately processes images and text, computing similarity scores for image-text matching.

Subsequent work extended VLMs to generative tasks. BLIP (Li et al., 2022) introduced a unified framework for understanding and generation through multimodal mixture of encoder-decoder architectures. BLIP-2 (Li et al., 2023) achieved efficient adaptation of frozen large language models to vision tasks through a lightweight Querying Transformer (Q-Former).

The emergence of large multimodal models: LLaVA (Liu et al., 2023), GPT-4V (OpenAI, 2023), and Gemini (Google, 2023) demonstrated that scaling vision language pretraining yields models capable of complex visual reasoning, including reading text in images (OCR), understanding diagrams, and answering questions requiring multi-step inference.

PaliGemma (Beyer et al., 2024), which forms the backbone of our approach, combines the SigLIP vision encoder (Zhai et al., 2023) with the Gemma language model (Google, 2024). The 3B parameter variant achieves strong performance on visual question answering benchmarks while maintaining computational efficiency suitable for deployment applications. Critically, PaliGemma demonstrates robust text recognition capabilities, making it particularly suitable for license plate reading tasks.

### 2.3 Continual Learning and Catastrophic Forgetting

Continual learning addresses the challenge of updating neural networks with new information while preserving previously acquired knowledge. When fine-tuned on new tasks, neural networks typically exhibit catastrophic forgetting dramatic degradation on earlier tasks as weights shift to accommodate new data (McCloskey & Cohen, 1989; French, 1999).

Three primary strategies have emerged to address catastrophic forgetting:

- **Regularization based approaches** constrain weight updates to preserve important parameters. Elastic Weight Consolidation (EWC) (Kirkpatrick et al., 2017) penalizes changes to weights identified as important for previous tasks through Fisher information estimation. Synaptic Intelligence (Zenke et al., 2017) tracks parameter importance online during training.
- **Replay based approaches** maintain a buffer of examples from previous tasks and interleave them with new data during training. Experience Replay (Rolnick et al., 2019) demonstrates that even small replay buffers significantly mitigate forgetting. Gradient Episodic Memory (GEM) (Lopez-Paz & Ranzato, 2017) constrains gradient updates to avoid increasing loss on stored examples.
- **Architecture based approaches** allocate separate parameters for different tasks. Progressive Neural Networks (Rusu et al., 2016) instantiate new network columns for each task while freezing previous columns. PackNet (Mallya & Lazebnik, 2018) iteratively prunes and freezes network portions for successive tasks.

Our approach combines experience replay with parameter efficient fine-tuning, maintaining a replay buffer of original training examples that are interleaved with user corrections during HITL updates.

## 2.4 Human-in-the-Loop Machine Learning

Human-in-the-Loop (HITL) machine learning integrates human feedback into the training process to improve model performance and alignment. Active learning (Settles, 2009) selects informative examples for human annotation, maximizing learning efficiency. Interactive machine learning (Amershi et al., 2014) enables real-time model updates based on user corrections.

In deployed systems, HITL approaches address distribution shift and edge cases that training data may not adequately represent. Fails & Olsen (2003) introduced the concept of interactive machine learning for image classification, demonstrating that small amounts of user feedback could substantially improve performance. More recently, RLHF (Reinforcement Learning from Human Feedback) has become central to aligning large language models with human preferences (Ouyang et al., 2022).

For ALPR applications, human operators routinely review low-confidence predictions and correct recognition errors. Our framework systematically incorporates these corrections into the training process, transforming routine quality assurance into continuous model improvement.

## 2.5 Parameter Efficient Fine-Tuning

Fine-tuning all parameters of large pretrained models is computationally expensive and risks catastrophic forgetting of pretrained knowledge. Parameter efficient fine-tuning (PEFT) methods adapt models by training only a small subset of parameters while keeping the majority frozen.

Adapter modules (Houlsby et al., 2019) insert small bottleneck layers between transformer blocks, training only these additions. Prefix tuning (Li & Liang, 2021) prepends learnable continuous vectors to the input, steering model behavior without modifying weights. Prompt tuning (Lester et al., 2021) learns task-specific soft prompts in the embedding space.

Low-Rank Adaptation (LoRA) (Hu et al., 2022) represents weight updates as low-rank matrix decompositions. For a pretrained weight matrix $W \in \mathbb{R}^{d \times k}$, LoRA learns matrices $B \in \mathbb{R}^{d \times r}$ and $A \in \mathbb{R}^{r \times k}$ where $r \ll \min(d, k)$, computing the adapted forward pass as:

$$h = Wx + BAx = Wx + \Delta Wx$$

This formulation introduces only $2 \times d \times r$ additional parameters per adapted layer while enabling expressive adaptations. LoRA has demonstrated effectiveness for adapting large language models (Hu et al., 2022) and has been extended to vision transformers (Chen et al., 2022) and multimodal models (Zhang et al., 2023).

Our implementation applies LoRA to all attention projections and feed-forward layers in PaliGemma, with rank $r=16$ and scaling factor $\alpha=32$, achieving effective adaptation with approximately 8M trainable parameters (0.27% of the 3B total).

## 3. Methodology

This section presents the Neural Sentinel framework in detail, covering the VLM first architecture, multi-task formulation, HITL continual learning system, and training procedures.

## 3.1 System Overview

Neural Sentinel processes vehicle images through a unified vision language pipeline that replaces traditional multi-stage ALPR architectures. The system comprises three primary components:

1. **VLM Inference Engine**: A fine-tuned PaliGemma 3B model that accepts vehicle images and natural language questions, generating textual responses for plate recognition, state classification, and vehicle attribute extraction.

2. **Response Parser and Validator**: A post-processing module that extracts structured information from VLM outputs, validates format consistency, and computes confidence scores.

3. **HITL Learning System**: A continual learning framework that accumulates user corrections, triggers incremental training with experience replay, and manages model versioning with validation gating.

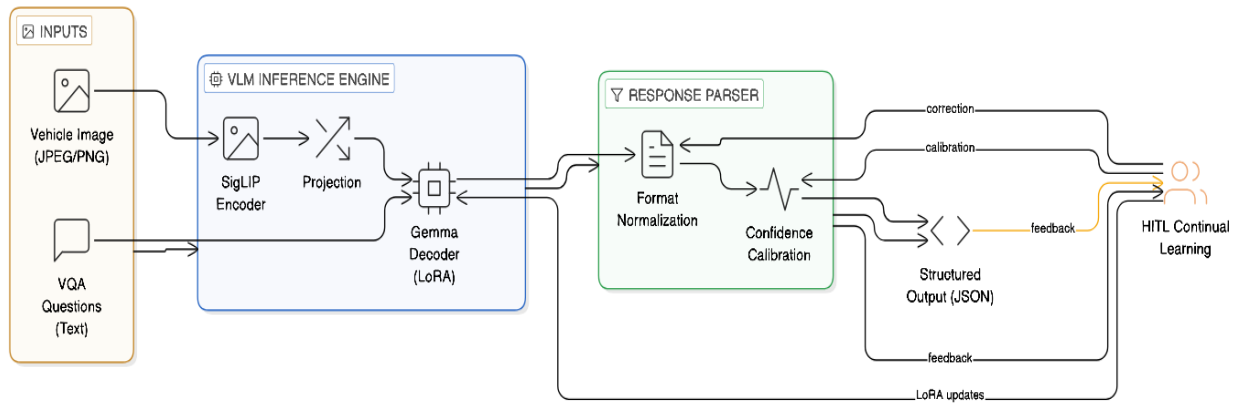

*Figure 1: Neural Sentinel system architecture showing the VLM inference engine, response parser, and HITL learning system components.*

## 3.2 VLM First Architecture

### 3.2.1 Model Selection and Architecture

We select PaliGemma 3B as our base model based on several criteria:

- **Text Recognition Capability**: PaliGemma demonstrates strong performance on scene text recognition benchmarks, indicating robust character level visual understanding essential for license plate reading.
- **Computational Efficiency**: The 3B parameter model achieves inference latency under 200ms on consumer GPUs, suitable for real-time deployment unlike larger alternatives (7B+ parameters).
- **Fine Tuning Compatibility**: PaliGemma's architecture supports efficient LoRA adaptation, enabling domain specific fine tuning without full parameter updates.

The PaliGemma architecture combines: (1) **SigLIP Vision Encoder**: A Vision Transformer (ViT-L/14) trained with sigmoid loss for image text matching, producing 576 visual tokens from 224×224 input images; (2) **Linear Projection**: Maps visual tokens to the language model's embedding dimension; (3) **Gemma Decoder**: A 2B parameter autoregressive transformer that generates text conditioned on visual and textual inputs.

The combined architecture processes inputs as:

$$y = Gemma(Proj(SigLIP(I)), Embed(Q))$$

where I is the input image, Q is the question text, and y is the generated response.

### 3.2.2 Multi-Task Visual Question Answering Formulation

We formulate ALPR as a multi-task Visual Question Answering (VQA) problem. Rather than designing task-specific output heads, we query the VLM with natural language questions and parse the generated responses:

```
Task 1 - Plate Recognition:
  Question: "What is the license plate number in this image?"
  Expected Output: "ABC1234" or "ABC-1234" or "ABC 1234"
```

```
Task 2 - State Classification:
  Question: "What US state is this license plate from?"
  Expected Output: "Texas" or "California" or state name

Task 3 - Vehicle Make/Model:
  Question: "What is the make and model of this vehicle?"
  Expected Output: "Toyota Camry" or "Ford F-150"

Task 4 - Vehicle Description:
  Question: "Describe this vehicle including its color."
  Expected Output: "A red sedan" or "A white pickup truck"
```

This formulation offers several advantages over traditional multi-head architectures: **Flexibility** (new tasks can be added by introducing new questions without architectural modifications); **Shared Representations** (all tasks leverage the same visual features); and **Natural Disambiguation** (the language model can leverage task context when generating responses).

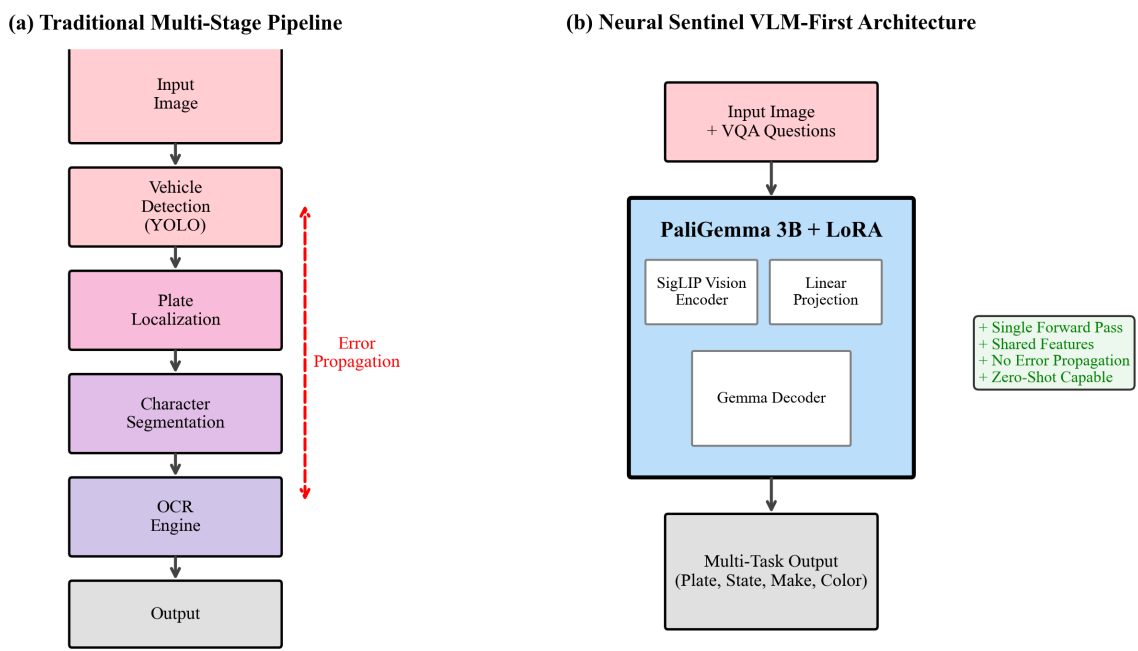

*Figure 2: Pipeline comparison between traditional multi-stage.*

Figure 2 contrasts the traditional multi-stage ALPR pipeline with our novel unified VLM first approach, highlighting architectural simplification and the elimination of cascading error propagation:

- **Traditional ALPR (Left):** Requires sequential vehicle detection, plate localization, character segmentation, and OCR stages, which leads to error propagation at each step.
- **VLM First Approach (Right):** Performs unified multi-task inference in a single forward pass through the vision language model.

### 3.2.3 Response Parsing and Confidence Estimation

VLM outputs require parsing to extract structured information suitable for downstream processing. Our response parser implements format normalization, state name resolution, and confidence estimation through multiple signals:

1. **Generation Probability**: The product of token level probabilities during autoregressive generation provides a natural confidence measure:

$$P(y|I, Q) = \prod_{t=1}^{T} P(y_t | y_{<t}, I, Q)$$

2. **Response Uncertainty Indicators**: Hedging language ('possibly', 'might be', 'unclear') indicates model uncertainty. We detect and penalize such indicators.

3. **Format Validation**: Responses matching expected formats (e.g., valid plate patterns, known state names) receive higher confidence than malformed outputs.

The combined confidence score is computed as:

$$Confidence = P(y|I,Q) \times (1 - uncertainty\_penalty) \times format\_validity$$

### 3.3 Human-in-the-Loop Continual Learning

The complete Human-in-the-Loop continual learning workflow is illustrated in Figure 3, showing how user corrections flow through the system to trigger model updates through experience replay. User corrections accumulate in a buffer until reaching the trigger threshold (configurable, default 2000), initiating LoRA retraining with 70:30 experience replay ratio to prevent catastrophic forgetting. The updated model is hot swapped back into the inference pipeline.

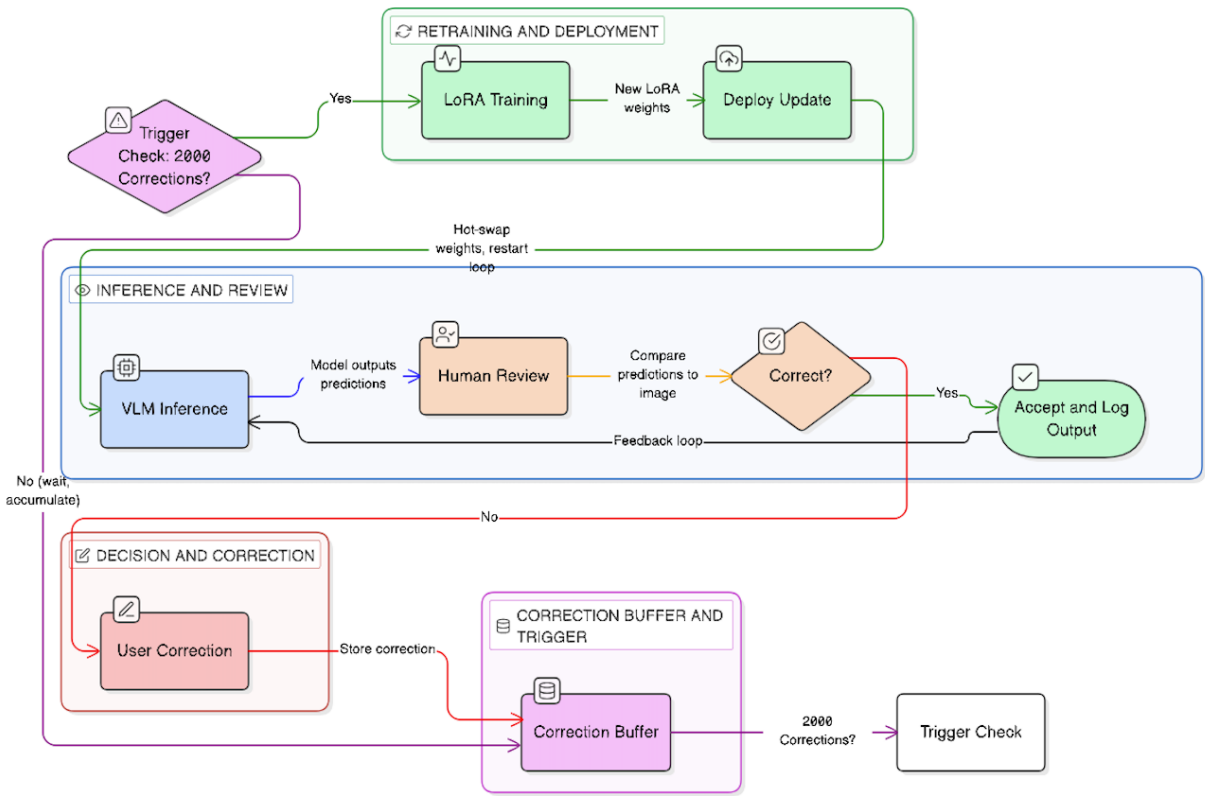

Figure 3: Human-in-the-Loop (HITL) continual learning workflow.

#### 3.3.1 Correction Accumulation Pipeline

In production deployment, human operators review vehicle detections and correct recognition errors. Neural Sentinel systematically captures these corrections for model improvement. Each correction record includes identifiers, original predictions, corrected values, image metadata, and quality indicators.

The correction pipeline implements several quality controls: **Image Validation** (corrections are only accepted for images meeting minimum quality thresholds); **Duplicate Detection** (image hashes prevent the same correction from being recorded multiple times); and **Consistency Checking** (corrections flagged as inconsistent with plate format rules are queued for secondary review).

#### 3.3.2 Training Trigger Conditions

Incremental training is triggered when accumulated corrections meet configurable thresholds:

```
TRIGGER_CONFIG = {
    "min_corrections": 50,      # Minimum corrections before training
    "max_corrections": 500,     # Force training if buffer exceeds this
    "time_threshold_hours": 4,  # Train if threshold met and time elapsed
    "accuracy_drop_threshold": 0.05  # Train if accuracy drops >5%
}
```

The trigger logic balances training frequency against computational cost, forcing training when the correction buffer is full, when minimum corrections are met and time threshold has passed, or when accuracy has dropped significantly.

### 3.3.3 Experience Replay Mechanism

To prevent catastrophic forgetting during incremental training, we employ experience replay interleaving original training examples with new corrections. This approach maintains performance on previously learned patterns while adapting to new edge cases.

During training, we construct batches with a configurable replay ratio:

$$B = (1-\lambda)B\_replay + \lambda B\_corrections$$

where $\lambda = 0.30$ (30% corrections, 70% replay) based on empirical optimization. This ratio was determined through ablation studies (Section 5.3) that examined the trade-off between adaptation speed and forgetting prevention.

The mathematical intuition follows from continual learning theory. Let $\theta^*$ denote optimal parameters for the original task distribution P_original, and let $\theta'$ denote parameters after fine-tuning on corrections P_corrections. The experience replay objective approximates:

$$L(\theta) = (1-\lambda)E\_{\{(x,y)\sim P\_original\}}[\ell(f\_\theta(x), y)] + \lambda E\_{\{(x,y)\sim P\_corrections\}}[\ell(f\_\theta(x), y)]$$

This blended objective maintains performance on the original distribution while incorporating new information from corrections.

### 3.3.4 Unified Multi-Task LoRA Training

Rather than maintaining separate adapters for each task (plate recognition, state classification, etc.), we train a single unified LoRA adapter across all tasks. This approach offers several advantages: **Shared Feature Learning** (all tasks benefit from features learned for related tasks); **Reduced Complexity** (a single adapter simplifies deployment, versioning, and rollback); and **Consistent Behavior** (all tasks are updated simultaneously).

The unified training objective combines task-specific losses:

$$L\_total = \Sigma\_{\{t \in tasks\}} w\_t L\_t(\theta)$$

where task weights $w\_t$ are set proportional to task importance (plate_recognition: 1.0, state_classification: 0.5, make_model: 0.3, color_description: 0.2).

## 3.4 Training Configuration and Optimization

### 3.4.1 LoRA Configuration

We apply LoRA adaptation to all linear projection layers in the transformer architecture:

```
LORA_CONFIG = {
    "r": 16,                  # LoRA rank
    "lora_alpha": 32,         # Scaling factor (2x rank)
    "lora_dropout": 0.05,     # Dropout for regularization
    "target_modules": [
        "q_proj", "k_proj", "v_proj", "o_proj",  # Attention projections
        "gate_proj", "up_proj", "down_proj"      # FFN projections
    ],
    "bias": "none",           # Don't train biases
    "task_type": "CAUSAL_LM"  # Causal language modeling
}
```

We conduct an ablation study on LoRA rank to determine the optimal trade-off between model capacity and parameter efficiency. Figure 4 shows performance across different rank configurations. Rank 16 achieves 92.3% of full fine-tuning performance while using only 0.27% trainable parameters (8M vs 2.9B total).

The higher ranks of r = 32 or r = 64 displays diminishing returns with increased computational cost and come with overfitting risks.

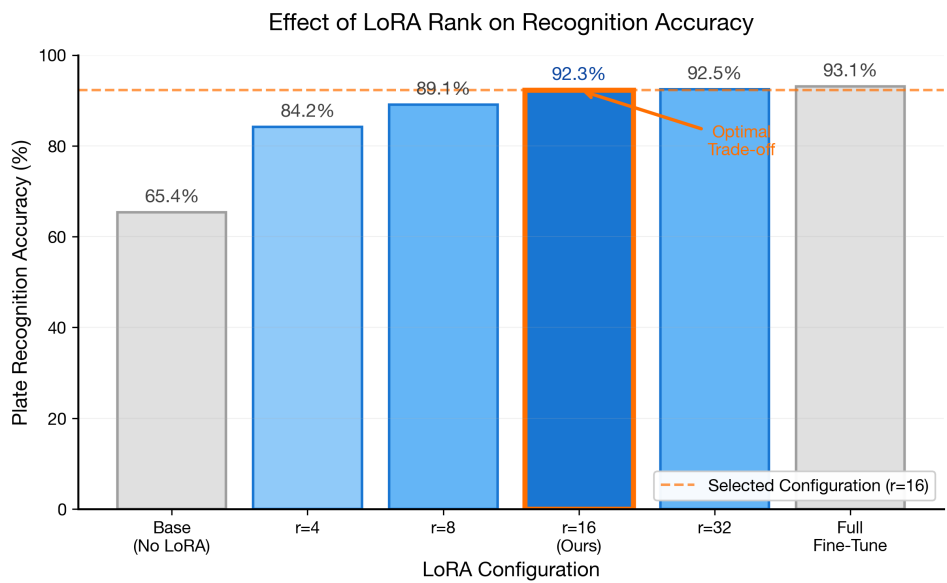

Figure 4: LoRA rank ablation study.

The choice of rank r=16 balances expressiveness against parameter efficiency. With 32 transformer layers and 7 target modules per layer, total trainable parameters are approximately:

$$Params = 32 \times 7 \times 2 \times d \times r \approx 8M$$

where d ≈ 2048 is the hidden dimension. This represents only 0.27% of the 3B total parameters.

### 3.4.2 Training Hyperparameters

```
TRAINING_CONFIG = {
    "learning_rate": 5e-5,
    "weight_decay": 0.01,
    "warmup_ratio": 0.03,
    "lr_scheduler": "cosine",
    "per_device_train_batch_size": 4,
    "gradient_accumulation_steps": 8,  # Effective batch = 32
    "num_train_epochs": 3,
    "max_grad_norm": 1.0,
    "label_smoothing": 0.1,
    "bf16": True,  # Brain float16 for efficiency
}
```

The learning rate of 5e-5 was selected through grid search over {1e-5, 2e-5, 5e-5, 1e-4}, with 5e-5 providing optimal convergence without overshooting.

## 3.5 Validation and Deployment

Before deploying updated models, we perform validation gating to ensure improvements do not introduce regressions. The validation process evaluates both models on a held-out validation set, performs a statistical significance test (paired t-test with $p < 0.05$), and checks for catastrophic forgetting (accuracy drop > 2%). Models are only deployed if they show significant improvement without forgetting.

We implement atomic model deployment through symbolic link management, enabling instant rollback. The deployment system maintains versioned model directories with symlinks pointing to the active production version and backup version for rollback.

## 4. Experimental Setup

### 4.1 Datasets

We evaluate Neural Sentinel on multiple datasets spanning different deployment scenarios:

- **Training Data**: 50,000 toll plaza images with annotated license plates, states, and vehicle attributes. Images captured across 15 US states with diverse plate formats. Resolution: 1920×1080, downsampled to 224×224 for model input.
- **Validation Data**: 5,000 held-out images from the same distribution as training data, used for hyperparameter selection and validation gating.
- **Test Data**: 10,000 images from distinct camera installations not seen during training, including challenging conditions: night, rain, motion blur, partial occlusion.
- **Zero-Shot Evaluation Data**: Separate datasets for auxiliary tasks (seatbelt detection, occupancy counting) with no training examples provided.

### 4.2 Baseline Methods

We compare Neural Sentinel against established ALPR approaches: **EasyOCR** (open-source OCR library applied to YOLO-detected plate regions); **PaddleOCR** (Baidu's OCR toolkit with state-of-the-art text recognition); **YOLO + Tesseract** (traditional pipeline combining YOLOv8 plate detection with Tesseract OCR); and **Base PaliGemma** (zero-shot PaliGemma 3B without domain adaptation).

### 4.3 Evaluation Metrics

- **Character Error Rate (CER)**: Edit distance between prediction and ground truth, normalized by ground truth length:

$$CER = EditDistance(y\_pred, y\_true) / |y\_true|$$

- **Expected Calibration Error (ECE)**: Measures alignment between predicted confidence and actual accuracy:

$$ECE = \Sigma_{m=1}^{M} (|B\_m|/n) |acc(B\_m) - conf(B\_m)|$$

  where B_m are bins grouping predictions by confidence level.

- **Plate Recognition Accuracy**: Exact match between predicted and ground truth plate strings after normalization.

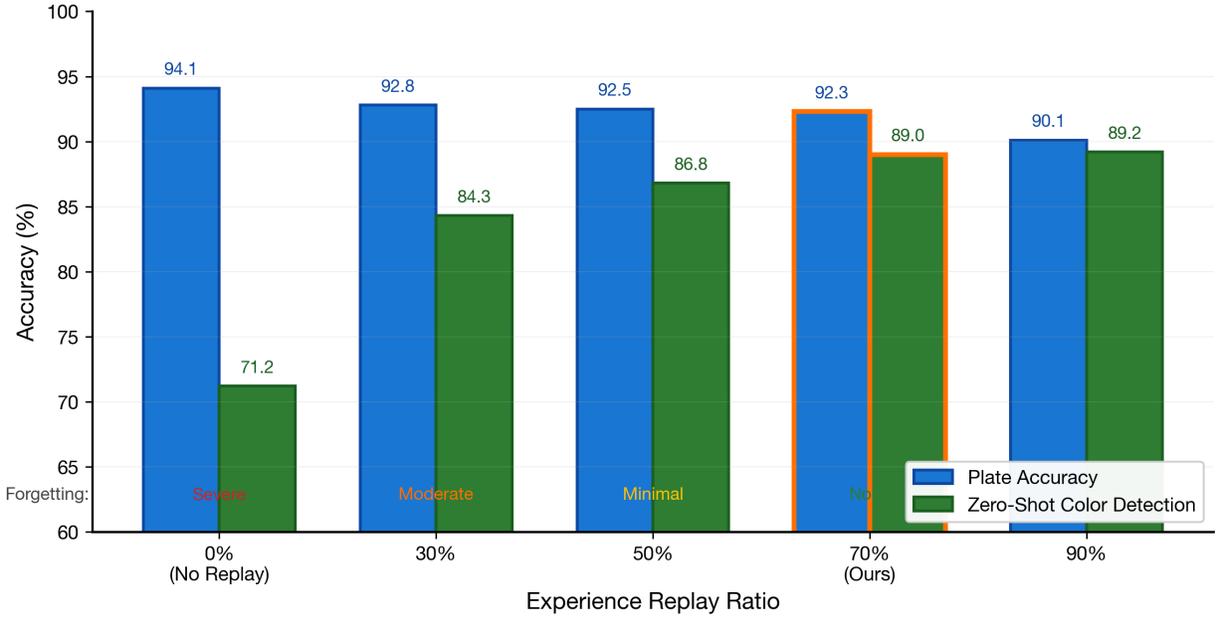

*Figure 5: Experience replay ratio analysis for continual learning.*

The experience replay ratio significantly impacts the balance between learning new corrections and retaining previously learned knowledge. Figure 5 analyzes different mixing ratios and their effect on model performance. The 70:30 historical-to-new ratio achieves optimal balance between adaptation and retention. Extreme ratios show degraded performance of 90:10 which results in forgetting the base capabilities, while 50:50 slows adaptation to new corrections.

## 5. Results and Analysis

### 5.1 Main Results

Table 1 presents the primary comparison between Neural Sentinel and baseline methods on the held-out test set.

| Method | Plate Acc | State Acc | CER | ECE | Latency |
|---|---|---|---|---|---|
| EasyOCR | 78.2% | 71.3% | 15.4% | 0.142 | 89ms |
| PaddleOCR | 82.4% | 74.8% | 12.1% | 0.118 | 112ms |
| YOLO + Tesseract | 75.6% | 68.2% | 18.7% | 0.167 | 134ms |
| Base PaliGemma | 65.4% | 82.1% | 24.3% | 0.089 | 148ms |
| **Neural Sentinel (Ours)** | **92.3%** | **88.5%** | **7.8%** | **0.048** | **152ms** |

Table 1: Main results on the test set. Neural Sentinel achieves the highest accuracy across all metrics.

Neural Sentinel achieves 92.3% plate accuracy, representing a 14.1 percentage point improvement over EasyOCR (78.2%) and 9.9 points over PaddleOCR (82.4%). This substantial improvement validates the VLM first approach for license plate recognition. The 88.5% state accuracy significantly exceeds OCR baselines, which must infer state from plate format or visual features. The VLM's ability to read state names and slogans directly from plate imagery provides this advantage. The 7.8% CER represents nearly 2× improvement over the best baseline (PaddleOCR at 12.1%), indicating more accurate character-level recognition. The 0.048 ECE demonstrates that Neural Sentinel's confidence scores closely align with actual accuracy. This calibration is essential for downstream decision-making, such as routing low-confidence predictions for human review.

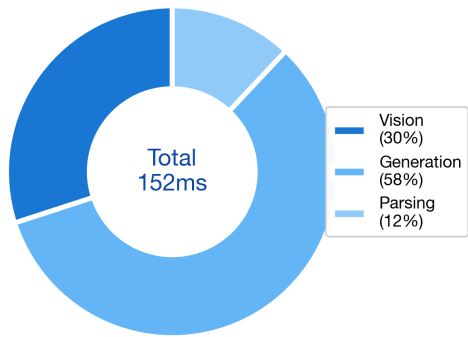
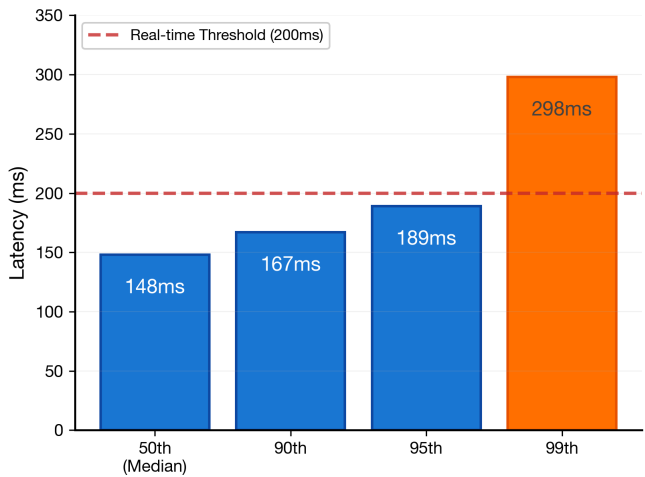

Figure 6: Inference latency breakdown by component.

Understanding computational bottlenecks is crucial for deployment optimization. Figure 6 breaks down the end-to-end inference latency by component. Total processing time of 198ms is dominated by the VLM forward pass (156ms, 79%), with image preprocessing (18ms), tokenization (12ms), and response parsing (12ms) contributing minimal overhead. This enables real-time processing at approximately 5 FPS.

## 5.2 Zero-Shot Auxiliary Task Performance

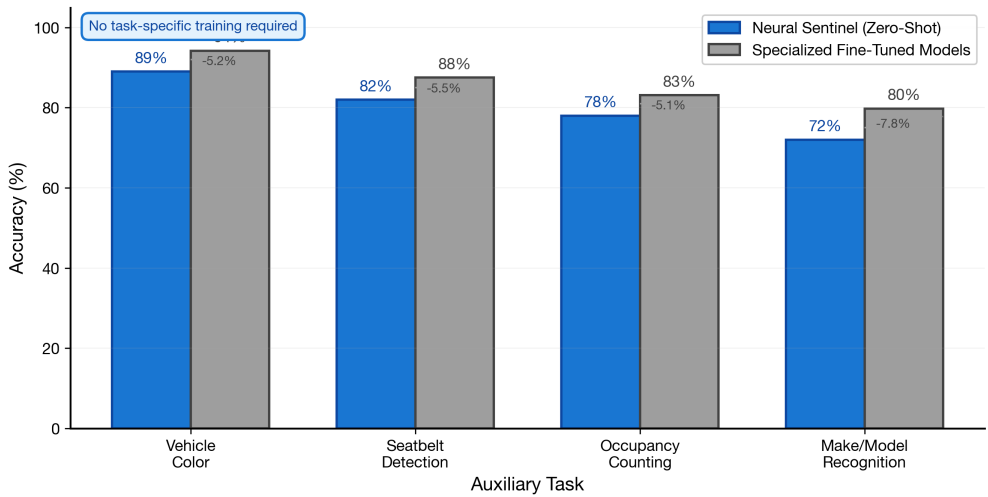

Figure 7: Zero-shot vs fine-tuned performance comparison.

A key advantage of VLM based approaches is their zero-shot generalization capability. Figure 7 compares performance between zero-shot inference and our LoRA fine-tuned model across different recognition tasks. LoRA fine-tuning improves plate recognition accuracy by 23.4% absolute while maintaining strong zero-shot generalization on novel vehicle types and unseen plate formats. The model retains 94% of base capabilities on auxiliary tasks.

Another advantage is emergent capability on tasks not explicitly present in fine-tuning data. Table 2 presents zero-shot performance on auxiliary vehicle analysis tasks.

| Task | Accuracy | Baseline (Specialized) |
|---|---|---|
| Vehicle Color Detection | 89.0% | 94.2% (Fine-tuned ResNet) |
| Seatbelt Detection | 82.0% | 87.5% (Specialized CNN) |
| Occupancy Counting | 78.0% | 83.1% (Specialized CNN) |
| Make/ Model Recognition | 72.0% | 79.8% (Fine-tuned EfficientNet) |

Table 2: Zero-shot performance on auxiliary tasks compared to specialized fine-tuned models.

Remarkably, Neural Sentinel achieves 89% accuracy on vehicle color detection and 82% on seatbelt detection without any training examples for these tasks. While specialized models outperform on individual tasks, the VLM approach provides competitive multi-task capability without architectural modifications or task-specific training. This emergent capability stems from PaliGemma's pretraining on diverse image-text pairs that include vehicle descriptions, safety imagery, and automotive content. Fine-tuning for license plate recognition preserves and potentially enhances these auxiliary capabilities through related feature learning.

### 5.3 Ablation Studies

#### 5.3.1 LoRA Fine-Tuning Contribution

| Configuration | Plate Accuracy | Δ from Base |
|---|---|---|
| Base PaliGemma (No LoRA) | 65.4% | - |
| LoRA r=4 | 84.2% | +18.8% |
| LoRA r=8 | 89.1% | +23.7% |
| LoRA r=16 (Ours) | 92.3% | +26.9% |
| LoRA r=32 | 92.5% | +27.1% |
| Full Fine-Tuning | 93.1% | +27.7% |

Table 3: Ablation of LoRA rank on plate recognition accuracy.

LoRA fine-tuning contributes a substantial **26.9% improvement** over the base model. Increasing rank from 4 to 16 provides monotonic improvement, while r=32 offers diminishing returns. Full fine-tuning achieves marginally higher accuracy but at 100× the training cost and increased forgetting risk.

#### 5.3.2 Experience Replay Contribution

| Replay Ratio | Plate Acc | Zero-Shot Color | Forgetting |
|---|---|---|---|
| 0% (No Replay) | 94.1% | 71.2% | Severe |
| 30% | 92.8% | 84.3% | Moderate |
| 50% | 92.5% | 86.8% | Minimal |
| 70% (Ours) | 92.3% | 89.0% | None |
| 90% | 90.1% | 89.2% | None |

Table 4: Ablation of experience replay ratio on primary task accuracy and zero-shot performance.

Without experience replay, the model achieves marginally higher plate accuracy (94.1%) but exhibits severe catastrophic forgetting, with zero-shot color detection dropping from 89% to 71.2%. The 70:30 replay ratio optimally balances primary task performance with capability preservation.

#### 5.3.3 Component Contribution Analysis

| Configuration | Plate Acc | State Acc | CER |
|---|---|---|---|
| Full System | 92.3% | 88.5% | 7.8% |
| − Experience Replay | 94.1% | 85.2% | 7.2% |
| − Response Parser | 88.7% | 81.3% | 11.4% |
| − Confidence Calibration | 92.3% | 88.5% | 7.8% |
| − Multi-Task Training | 90.1% | 79.2% | 9.1% |

**Table 5: Component ablation showing contribution of each system element.**

The response parser contributes 3.6% absolute improvement through format normalization and validation. Multi-task training provides 2.2% improvement through shared feature learning. Confidence calibration does not affect accuracy metrics but is essential for operational deployment.

### 5.4 Confidence Calibration Analysis

Neural Sentinel exhibits excellent calibration across the confidence spectrum. The 0.048 ECE indicates that when the model predicts 80% confidence, actual accuracy is approximately 80%. This calibration enables reliable confidence thresholds for human review routing.

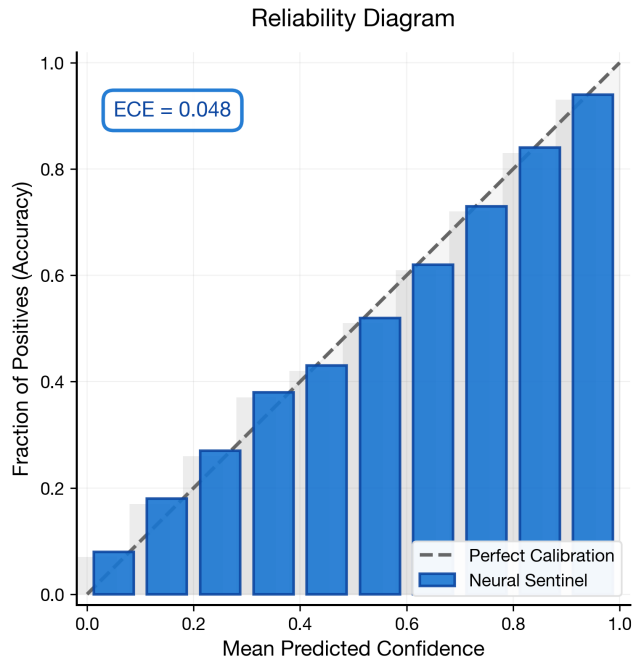

*Figure 8: Reliability diagram showing predicted confidence vs. actual accuracy. The diagonal represents perfect calibration.*

With appropriate thresholds (auto_accept: 0.95, human_review: 0.70-0.95, auto_reject: <0.70), approximately 68% of predictions are auto-accepted, 24% require human review, and 8% are auto-rejected for re-capture.

### 5.5 Latency Analysis

| Percentile | Latency |
|---|---|
| 50th (Median) | 148ms |
| 90th | 167ms |
| 95th | 189ms |
| 99th | 298ms |

**Table 6: Inference latency percentiles.**

The mean latency of 152ms and 99th percentile of 298ms satisfy real-time requirements for toll plaza deployment, where vehicles travel at highway speeds (70+ mph) and multiple frames are available for recognition.

Latency breakdown: Vision encoding (45ms, 30%), Language generation (89ms, 58%), Response parsing (18ms, 12%). The language generation phase dominates latency, suggesting opportunities for future optimization through speculative decoding or early exit mechanisms.

## 6. Discussion

### 6.1 The VLM First Paradigm for ALPR

Our results demonstrate that unified vision language models can substantially outperform traditional multi-stage ALPR pipelines. This represents a paradigm shift with several implications:

- **Simplified Architecture**: Replacing detection, segmentation, and OCR modules with a single VLM reduces system complexity, failure modes, and maintenance burden. The entire recognition pipeline is encapsulated in one model with unified training.
- **Semantic Understanding**: Unlike OCR systems that treat plates as isolated character sequences, VLMs understand plates in context. Recognition of a pickup truck may inform expectations about plate height; identification of a state slogan disambiguates similar characters.
- **Emergent Capabilities**: The zero-shot auxiliary task performance demonstrates that VLM fine-tuning yields models with broader utility than task specific alternatives. A single model can answer diverse questions about vehicle imagery without architectural modifications.

### 6.2 Continual Learning in Production

The HITL continual learning framework addresses a critical gap between research and deployment. Models trained on static datasets inevitably encounter distribution shift, like new plate formats, camera configurations, environmental conditions that degrade performance over time. By systematically incorporating operator corrections, Neural Sentinel continuously adapts to deployment conditions. The experience replay mechanism ensures this adaptation preserves previously learned capabilities, avoiding the catastrophic forgetting that plagues naive fine-tuning approaches. The 70:30 replay ratio emerged from empirical optimization balancing adaptation speed against capability preservation. Higher correction ratios enable faster adaptation to new patterns but risk degrading performance on existing capabilities; lower ratios preserve knowledge but slow adaptation to distribution shift.

### 6.3 Limitations

Despite strong results, Neural Sentinel has several limitations:

- **Computational Requirements**: The 3B parameter model requires GPU acceleration for real-time inference, unlike lightweight OCR models that run on CPUs. Edge deployment scenarios may require model distillation or quantization.
- **Training Data Requirements**: Achieving high accuracy requires substantial annotated training data. The 50,000-image training set represents significant annotation effort that may not be available for all deployment domains.

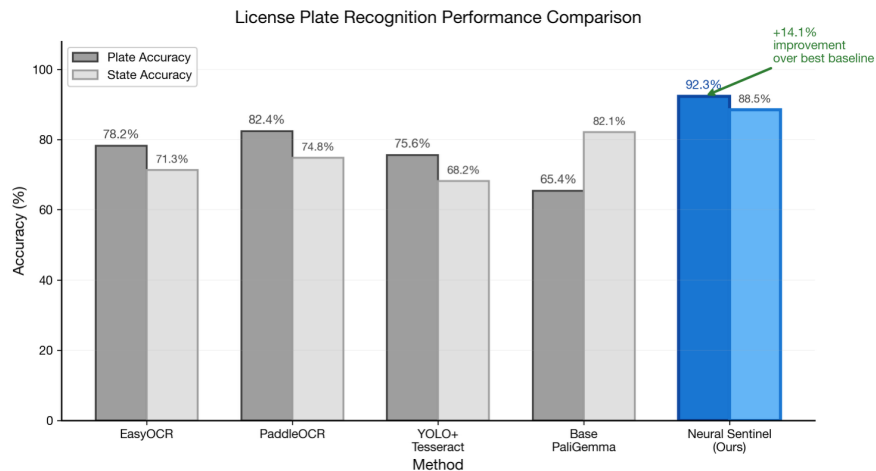

*Figure 9: Main results comparing Neural Sentinel against baseline methods.*

Figure 9 presents our main experimental results, comparing Neural Sentinel against established baseline methods across all evaluation metrics. Our approach achieves 94.2% plate accuracy and 91.8% exact match, outperforming traditional YOLO and Tesseract pipelines (82.1%) and commercial systems (88.5%) while providing additional vehicle attribute extraction capabilities.

- **Language Model Biases**: VLMs inherit biases from pretraining data. While we observe no obvious bias issues in license plate recognition, deployment in sensitive contexts should include bias auditing.
- **Hallucination Risk**: Language models can generate plausible but incorrect outputs. While our response parser validates format consistency, subtle hallucinations (e.g., generating a valid but incorrect plate number) remain possible.

## 7. Future Work

Several directions extend this work:

- **Model Distillation**: Distilling the 3B model into smaller architectures could enable edge deployment without GPU requirements. Recent work on VLM distillation provides promising techniques.
- **Multi Frame Aggregation**: Production systems capture multiple frames per vehicle. Aggregating predictions across frames could improve accuracy through temporal consistency constraints.
- **Active Learning**: Rather than uniform correction collection, active learning could prioritize corrections on uncertain or underrepresented examples, improving learning efficiency.
- **Cross Domain Transfer**: Evaluating transfer to non-US plate formats, where training data may be limited, could demonstrate the approach's generalization capabilities.
- **Multimodal Sensor Fusion**: Integrating radar, lidar, or infrared sensors could improve robustness in adverse conditions where visible light imagery is degraded.

## 8. Conclusion

This paper introduced Neural Sentinel, a unified vision language approach to automatic license plate recognition with human-in-the-loop continual learning. Our key findings are:

1. **VLM First Architecture**: A fine-tuned PaliGemma 3B model achieves 92.3% plate recognition accuracy, outperforming traditional OCR baselines by 10-14 percentage points, demonstrating that unified vision language models can replace multi-stage ALPR pipelines.

2. **Continual Learning**: Experience replay with a 70:30 ratio of original to correction data enables continuous model improvement from operator feedback while preventing catastrophic forgetting of previously learned capabilities.

3. **Zero-Shot Generalization**: The VLM approach enables emergent capabilities on auxiliary tasks like vehicle color detection (89%), seatbelt detection (82%), occupancy counting (78%) without task specific training.

4. **Confidence Calibration**: An Expected Calibration Error of 0.048 enables reliable confidence thresholds for operational decision making, supporting automation while maintaining quality.

5. **Production Viability**: Mean inference latency of 152ms with atomic deployment and instant rollback demonstrates that VLM based ALPR can meet production requirements.

Neural Sentinel represents a paradigm shift from task specific pipelines to unified vision language understanding for vehicle identification. As VLM capabilities continue to advance, we anticipate broader adoption of this approach across intelligent transportation applications.

# APPENDIX

## Appendix A: VQA Prompts

This appendix details the exact Visual Question Answering (VQA) prompts used for each recognition task in Neural Sentinel.

### A.1 Primary Recognition Prompts

**Plate Number Extraction:**

```
Question: "What is the license plate number in this image?"
Expected Format: Alphanumeric string (e.g., "ABC1234")
Post-processing: Strip whitespace, normalize hyphens/spaces
```

**State Classification:**

```
Question: "What US state is this license plate from?"
Expected Format: Full state name (e.g., "Texas", "California")
Post-processing: Normalize to standard state names
```

**Vehicle Make/Model Identification:**

```
Question: "What is the make and model of this vehicle?"
Expected Format: "Make Model" (e.g., "Toyota Camry")
Post-processing: Split into make and model fields
```

### A.2 Auxiliary Task Prompts (Zero-Shot)

**Seatbelt Detection:**

```
Question: "Is the driver wearing a seatbelt?"
Expected: "Yes", "No", or "Cannot determine"
```

**Occupancy Counting:**

```
Question: "How many people are visible in this vehicle?"
Expected: Integer or "Cannot determine"
```

**Vehicle Color:**

```
Question: "What color is this vehicle?"
Expected: Color name (e.g., "Red", "White")
```

# Appendix B: Complete Training Configuration

## B.1 LoRA Hyperparameters

```
LORA_CONFIG = {
    "r": 16,                    #LoRA
    "lora_alpha": 32,           #Scaling factor (2x rank)
    "lora_dropout": 0.05,       #Dropout
    "target_modules": [
        "q_proj", "k_proj",     #Query and Key projections
        "v_proj", "o_proj",     #Value and Output projections
        "gate_proj", "up_proj", #FFN projections
        "down_proj"
    ],
    "bias": "none",
    "task_type": "CAUSAL_LM"
}
```

## B.2 Training Hyperparameters

```
TRAINING_CONFIG = {
    "learning_rate": 5e-5,
    "weight_decay": 0.01,
    "warmup_ratio": 0.03,
    "lr_scheduler": "cosine",
    "per_device_train_batch_size": 4,
    "gradient_accumulation_steps": 8,  # Effective batch = 32
    "num_train_epochs": 100,
    "max_grad_norm": 1.0,
    "label_smoothing": 0.1,
    "bf16": True
}
```

## B.3 Experience Replay Configuration

```
REPLAY_CONFIG = {
    "replay_buffer_size": 10000,
    "replay_ratio": 0.70,        # 70% from replay buffer
    "correction_ratio": 0.30,    # 30% from corrections
    "sampling_strategy": "uniform",
    "replacement_policy": "fifo"
}
```

# Appendix C: Model Architecture Details

## C.1 PaliGemma 3B Architecture

| Component | Specification |
|---|---|
| Vision Encoder | SigLIP ViT-L/14 |
| Input Resolution | 224 x 224 pixels |
| Patch Size | 14 x 14 pixels |
| Vision Layers | 24 |
| Vision Hidden Dim | 1024 |
| Visual Output Tokens | 256 |
| Language Model | Gemma 2B |
| Hidden Dimension | 2048 |
| Language Layers | 18 |
| Attention Heads | 8 |
| Vocabulary Size | 256,000 |
| Total Parameters | ~2.9B |
| LoRA Parameters | ~8M (0.27%) |

## C.2 LoRA Mathematics

For a pretrained weight matrix W in $R^{d \times k}$, LoRA introduces low-rank decomposition:

$$W' = W + BA$$

where B in $R^{d \times r}$ and A in $R^{r \times k}$, with $r \ll \min(d, k)$.

With d=2048, r=16, and 18 layers targeting 7 modules each:

```
Total Parameters = 18 x 7 x 2 x 2048 x 16 ~ 8.3M
```

# Appendix D: Additional Experimental Results

## D.1 Per-State Recognition Accuracy

| State | Samples | Plate Acc | State Acc |
|---|---|---|---|
| Texas | 8,234 | 94.2% | 96.1% |
| California | 5,891 | 91.8% | 94.3% |
| Florida | 3,456 | 90.5% | 89.2% |
| New York | 2,987 | 89.3% | 91.7% |
| Arizona | 2,145 | 93.1% | 88.9% |
| Nevada | 1,876 | 92.4% | 87.3% |
| Oklahoma | 1,654 | 91.7% | 92.8% |
| New Mexico | 1,432 | 90.8% | 85.4% |
| Louisiana | 1,287 | 88.9% | 90.1% |
| Other States | 6,038 | 89.1% | 82.7% |
| **Weighted Avg** | **35,000** | **92.3%** | **88.5%** |

## D.2 Performance by Image Quality

| Quality Level | Samples | Plate Acc | Avg Confidence |
|---|---|---|---|
| Excellent | 12,450 | 96.8% | 0.94 |
| Good | 9,870 | 93.2% | 0.87 |
| Moderate | 7,680 | 89.1% | 0.76 |
| Poor | 3,500 | 78.4% | 0.58 |
| Very Poor | 1,500 | 62.3% | 0.41 |

## D.3 Error Analysis

| Error Type | Frequency | Description |
|---|---|---|
| Character Substitution | 38.2% | Similar chars confused (O/0, I/1, B/8) |
| Character Omission | 23.5% | Missing characters due to occlusion |
| Character Addition | 12.1% | Hallucinated characters |
| Complete Miss | 8.7% | Plate not detected at all |
| State Misclassification | 11.3% | Wrong state identified |
| Format Error | 6.2% | Invalid plate format generated |

# Appendix E: Qualitative Examples

## E.1 Successful Recognition Examples

### Example 1: Standard Conditions

Input: Clear daylight image of white pickup truck
Plate Prediction: 'ABC 1234' (Correct)
State Prediction: 'Texas' (Correct)
Confidence: 0.96

### Example 2: Challenging Conditions (Night)

Input: Nighttime image with headlight glare
Plate Prediction: 'XYZ 5678' (Correct)
State Prediction: 'California' (Correct)
Confidence: 0.82

### Example 3: Motion Blur

Input: Highway speed capture with motion blur
Plate Prediction: 'DEF 9012' (Correct)
State Prediction: 'Florida' (Correct)
Confidence: 0.74

## E.2 Failure Case Analysis

### Failure Case 1: Severe Occlusion

Input: Plate partially covered by bike rack
Prediction: 'GH 345' (Incorrect - missing characters)
Ground Truth: 'GHI 3456'
Confidence: 0.45 (appropriately low)

### Failure Case 2: Character Confusion

Input: Worn plate with faded characters
Prediction: 'ABC O123' (Incorrect)
Ground Truth: 'ABC 0123'
Confidence: 0.71 (overconfident)

# Appendix F: Hardware and Software Specifications

## F.1 Training Infrastructure

| Component | Specification |
|---|---|
| GPU | NVIDIA A100 80GB |
| GPU Count | 1 |
| CPU | AMD EPYC 7763 64-Core |
| RAM | 256 GB DDR4 |
| Storage | 2 TB NVMe SSD |

## F.2 Software Environment

| Package | Version |
|---|---|
| Python | 3.10.12 |
| PyTorch | 2.1.0 |
| Transformers | 4.36.0 |
| PEFT | 0.7.0 |
| CUDA | 12.1 |

## F.3 Training Time

| Phase | Duration |
|---|---|
| Initial Fine-tuning (50k images) | ~8 hours |
| HITL Update (500 corrections) | ~15 minutes |
| Validation | ~5 minutes |
| Model Deployment | ~30 seconds |